# Multi-view Gait Recognition based on Siamese Vision Transformer

Yanchen Yang, Ruoyu Li,
College of Information, Yunnan Normal University, China.
yangyanchen@user.ynnu.edu.cn
liruoyu@user.ynnu.edu.cn

Lijun Yun*, Feiyan Cheng,
College of Information, Yunnan Normal University, China.
Yunnan Key Laboratory of Optoelectronic Information Technology, China.
yunlijun@ynnu.edu.cn
chengfy03@163.com

***Abstract:*** While the Vision Transformer has been used in gait recognition, its application in multi-view gait recognition is still limited. Different views significantly affect the extraction and identification accuracy of the characteristics of gait contour. To address this, this paper proposes a Siamese Mobile Vision Transformer (SMViT). This model not only focuses on the local characteristics of the human gait space but also considers the characteristics of long-distance attention associations, which can extract multi-dimensional step status characteristics. In addition, it describes how different perspectives affect gait characteristics and generate reliable perspective feature relationship factors. The average recognition rate of SMViT on the CASIA B data set reached 96.4%. The experimental results show that SMViT can attain state-of-the-art performance compared to advanced step recognition models such as GaitGAN, Multi_view GAN, Posegait and other gait recognition models.

# 1 Introduction

Human walking posture, another emerging characteristic of biological features such as facial, fingerprints, and iris, has the advantages of long-distance, non-contact, and inductive identification[1]. It has occurred in terms of identity identification. In addition, as the security facilities of urban and public places are gradually improved, monitoring facilities such as cameras are everywhere so that we can use uncomplicated low-resolution instruments without the perception of others [2]. Character characteristics to judge their identity. It also makes the gait recognition method based on deep learning in today's society, which has a broader application prospect [3], especially in the criminal investigation of public security, and has an amazing advantage [4]. Thus, the research on gait recognition is of great significance to the long-term steadiness of national society.

Regardless, if you want a sense of identification with pedestrians in public places, you must overwhelm the randomness of pedestrians walking in public places. Collecting and identifying pedestrian gait information from multiple views [5,6]. Formal gait recognition uses 90 ° gait features that show the most considerable vivid details about human posture as experimental data. The rationale is that the gait characteristics in other views overlap due to the perspective problems of the human parts, and the contour characteristics cannot be effectively shown. This is also one of the complications of multi-view gait recognition. In addition, in the actual application scenario, to retain the advantage of inductive identification, we cannot fix the pedestrian walking position and relative view with the camera [7]. The problem needs to be solved urgently.

In the multi-view gait recognition task, when the view is shifted from 90 ° to 0 ° and 180 °, it will be gone with the view so that when the outline of the walking posture is extracted, the degree of coincidence of the body part is U-shaped. It earnestly affects the consequence of extraction of gait contour characteristics. In response to this issue, this paper uses Siamese neural network to calculate the posture relationship between the two views of the two views and calculate the characteristic conversion factor. Under the premise of retaining identity information, the helpful high-dimensional intensive characteristics are strengthened to make its high-dimensional features more vivid, and the effect of loss of gait characteristics on recognition accuracy is weakened. The SMViT model constructed through this concept can obtain higher recognition accuracy in a multi-view of non 90 °, and the model is more robust.

In summary, this paper makes the following contributions:

(1)Design a reasonable and novel gait view conversion method, which can deal with the problem of multi-view gait.

(2)Construct the SMViT model, and use the view characteristic relation module to calculate the gait characteristics association between multi-view.

(3) We discovered a gradually moving view training strategy that can raise the model's robustness while raising the recognition rate for less precise gait view data.

The structure of this paper is as follows: The related technologies of gait recognition will be introduced in Sect. 2. Then the SMViT constructed in this paper and the view gradually moving training method are explained in Sect. 3. In Sect. 4, use the CASIA B gait dataset [8] experimentation and explore the models and methods presented in this paper. Finally, in Sect. 5, we will summarize the full text and look forward to the future direction of gait recognition.

## 2 Related work

At present, there are many ways to solve the problem of multi-view gait recognition. Some researchers adopt the method of constructing three-dimensional model, and use the close cooperation of multiple cameras to construct the three-dimensional model of pedestrian movement, so as to weaken the influence of perspective, clothing and other factors. Bodor R et al. proposed to combine arbitrary views taken by multiple cameras to construct appropriate images to match the training view for pedestrian gait recognition [9]. Ariyanto G et al. constructed the correlation energy map between their proposed generative gait model and the data, and adopted the dynamic programming method to select possible paths and extract gait kinematic trajectory features, and believed that the extracted features were inherent to 3D data [10]. In addition, Tome D et al. constructed a comprehensive method to fuse the probability knowledge of 3D human pose with CNN, and proposed a unified formula to solve the problem of estimating 3D human pose from a single original RGB image [11]. Later, Weng J et al. changed the extraction method of human 3D pose and proposed a deformable pose ergodic convolution to optimize the convolution kernel of each joint by considering context joints with different weights [12]. However, this method of 3D pose modeling is more complicated to calculate, and has high requirements on the number of cameras and shooting environment, so it is difficult to be applied in the application scene with only ordinary cameras.

Some scholars used the View Transformation Model (VTM) to extract the frequency domain features of gait contours by transforming their views from different views. Among them, Makihara Y et al., proposed a gait recognition method based on frequency domain features and view transformation. Firstly, the spatio-temporal contour set of gait was constructed, and the periodic characteristics of gait were used for Fourier analysis to extract the frequency domain features of pedestrian gait, and the multi-view training set was used to calculate the view transformation model [13]. In this method, the spatiotemporal gait images in the gait cycle are usually first fused into the Gait Energy Image (GEI), which is a spatiotemporal gait representation method proposed by Han J et al. [14]. Kusakunniran W et al. combined Gait Energy Image (GEI) with view transition model (VTM), and used Linear Discriminant Analysis (LDA) to optimize the feature vectors and improve the performance of VTM [15]. Later, Kusakunniran W et al. used a motion clustering method to classify gaits from different views into a group according to their correlation, and within each group, canonical correlation analysis (CCA) was used to further enhance the linear correlation between gaits from different views [16]. In addition, the researchers focused on gait recognition from any view, Hu M et al. proposed a Viewpoint Invariant Discriminant Projection (ViDP) method to improve the discrimination accuracy of gait features by linear projection [17]. However, most of these methods are realized by domain transformation or singular value decomposition, and the perspective of transformation is complicated.

Others have used an adversarial generative network to normalize multiple views into a common perspective. Zhang P et al. proposed a perspective-shifting Adversarial Generative Network (VT-GAN), which can transform gait views across two arbitrary views with only one model [18]. Shi y et al. designed GaitGANv1 and GaitGANv2, a gait adversarial generation network, which used GAN as a regressor to generate a standardized side view of normal gait, not only to prevent falsification of gait images, but also to maintain identity information, and achieved good results in the cross-view gait recognition task [19,20]. In addition, Wen J et al. used GAN to convert gait images with arbitrary decorations and views into normal states of 54°, 90° and 126°, so as to extract view-invariant features and reduce the loss of feature information caused by view transformation [21]. Aiming at the problem of limited recognition accuracy due to the lack of gait samples from different views, Chen X et al. proposed a Multi-View Gait Generation Adhoc Network (MvGGAN) to generate false gait samples to expand the dataset and improve the recognition accuracy [22]. However, the recognition accuracy of this adversarial generative network structure in the same perspective is easily affected by decorative features such as clothes and backpacks, resulting in limited recognition accuracy.

# 3 SMViT and view gradually moving training method

In order to solve the multi-perspective situation, this paper uses the Siamese neural network as the design idea, calculates the correlation between the characteristics of different views, and uses this as the basis for the conversion of the characteristics of the view. When there are few specimens, a Siamese neural network can extract and learn the link between two groups of photos [23]. ViT has an advantage for extracting multi-scale features because of its robust strength and resistance to interference from mistake samples [24,25].In this paper, a two-channel Siamese module (Conv and MViT, CM Block) of convolution is constructed to extract characteristics of multi-view gait contour features. The specific model structure is shown in Figure 1.

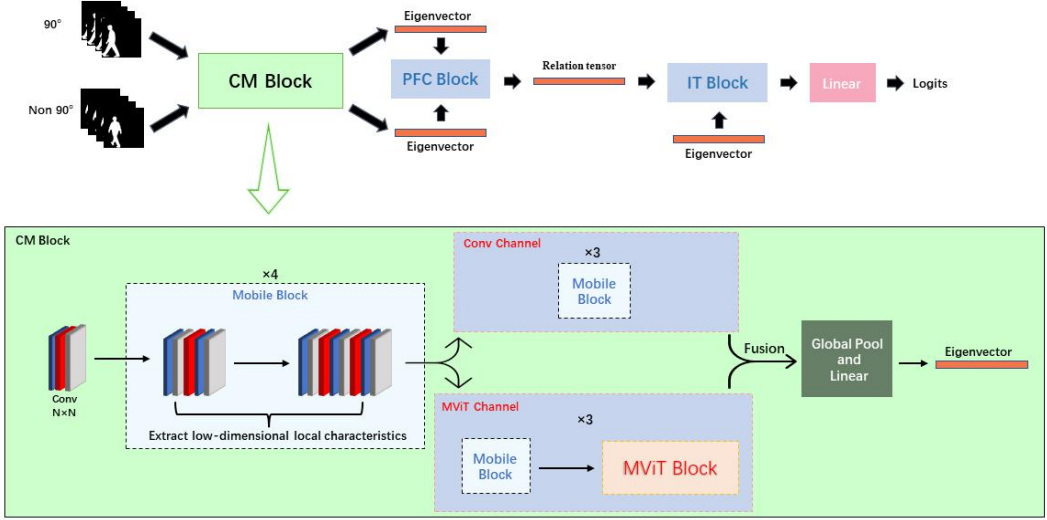

**Figure 1**: Structure diagram of SMViT

In order to extract gait information from various viewpoints, two feature extraction networks are used in the Siamese network module in this paper. Convolution channels are used inside each module to obtain the contour's high-dimensional local features. Furthermore, we utilize the Mobile ViT channel to create the high-dimensional states that are indicative long-distance attention characteristics of the current view. The Inverse Transformation block (IT block) and the Perspective feature conversion block (PFC block) are designed concurrently. The former is used to calculate the characteristics of the two perspectives obtained by the Siamese network, and the relation tensor is taken as the view conversion factor, as shown in Formula (1). The latter is used to convert the characteristics of high-dimensional between two views, as shown in the formula (2). Among them, $x$ and $y$ are two view cornering characteristics, respectively, and $N$ is the capacity of the target view set.

$$PFC(x,y) = \frac{\sum_{i=1}^{N}(x_i - y_i)}{N} \qquad (1)$$

$$IT(x,y) = x + PFC(x,y) \qquad (2)$$

In addition, use the Mobile View Transformation (MVT) module to extract the view characteristics and tensor. This module is based on the Mobile ViT model to incorporate the convolution with Transformer. Not losing the order of the patch nor affecting the inner space relation of the pixels in each patch. Replace the local processing in the convolution with deeper global treatment. In an effective receiving domain, we model long-distance non-local dependencies. Come to retain the advantages of convolution and ViT in the extraction of gait contour features. Has a smaller parameter and ideal experimental results. The specific details of the module are shown in Figure 2.

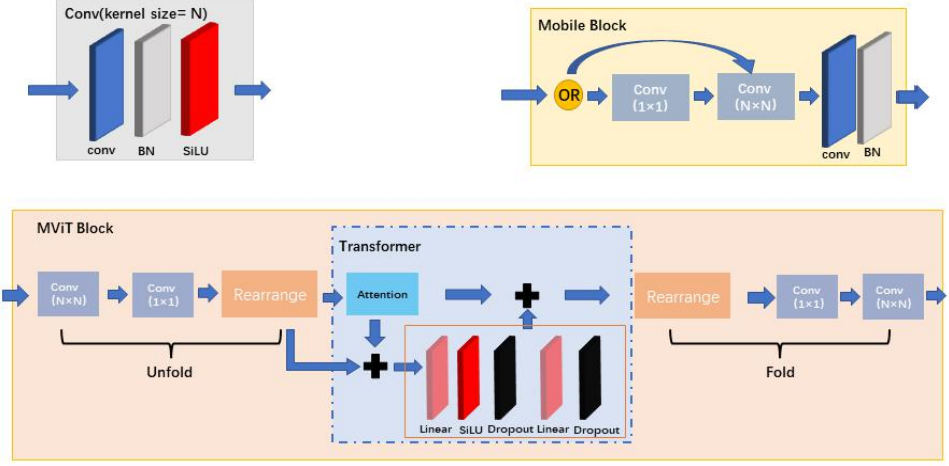

**Figure 2**: Details of the module

In Figure 2, the Conv block module (Conv Block) is divided into two types: $N \times N$ convolution and point-by-point convolution according to the difference in convolution kernel size. This module consists of a convolution layer and a Batch Normalization layer. Mobile Block takes MobileNet as the essential train of thought [26-28] and controls the network depth and the number of parameters by constructing depth-separable convolution. In addition, due to the distinction in the processing method and content of feature extraction between convolution and ViT, format conversion needs to be carried out before and after the Transformer module to control the data processing format. SiLU is used as the activation function in each module, as shown in Equation (3), and the global average pooling layer is used in the pooling layer, as shown in Equation (4), Where $x$ means the input matrix, and $x_w$ represents the operation area of the pooling layer.

$$SiLU(x) = x \cdot Sigmoid(x) \quad (3)$$

$$Pooling(x_w) = Avg(x_w) \quad (4)$$

The Transformer module absorbs some of the advantages of convolutional computing and retains its characteristic processing capabilities in the space perception domain. By dividing a large global receptive field into different patches in a non-overlapping way, $P = Wh$ where $w$ and $h$ are the width and height of patches respectively, and then Transformer is used to encode the relationship between patches. Among them, the self-attention mechanism calculates the scaled dot product attention by constructing the query vector Q, the value vector V and the key vector K, as shown in Equations (5), (6) and (7).

$$f(Q,K) = \frac{QK^T}{\sqrt{d_k}} \quad (5)$$

$$X = softmax(f(Q,K)) \quad (6)$$

$$Attention(X,V) = X \times V \quad (7)$$

In this case, the module's computation cost of multi-head self-attention is $O(N^2 Pd)$. Compared with the traditional ViT, the calculation cost $O(N^2 d)$ is increased, but the speed is faster in practical applications [29].

# 4 Experiment and Analysis

**4.1 Experimental Data**

CASIA B dataset is a large dataset widely used in multi-view gait recognition tasks. It consisted of 124 subjects (31 women and 93 men) [19], In 11 sets of view from 0° to 180° (with an interval of 18° for close views), Gait images of each subject in three different states of normal walking (NM), walking with bag (BG) and walking with coat (CL) were collected [30], as shown in Figure 3.

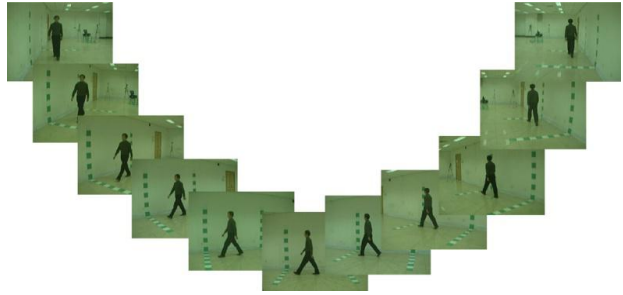

**Figure 3**: CASIA B multi-view gait dataset

**4.2 Experimental Design**

In the multi-view gait recognition task, due to the offset of the perspective, part of human gait contour characteristics are lost and the accuracy is reduced. This experiment designed 90 ° as a high-precision standard perspective. The remaining 10 views of the gait characteristics are calculated by the standard view relationship. Don't consider the mutual verification between the perspective, only the gait recognition accuracy inside the perspective is calculated. And establish a simple perspective conversion factor group to transform the view feature tensor with less feature information, To the 90° feature tensor with more prosperous and distinct feature information. The training set and verification set in the experiment are divided according to the ratio of 7: 3. There is no crossing between each view, which is mainly aimed at improving the recognition accuracy within each view. In addition, the gait data obtained in the actual application scenario may not necessarily contain a complete gait cycle, and gait characteristics are random. Therefore, this experiment does not adopt the gait cycle group as data input. Instead, the gait group with three attitudes is scattered at will to ensure that the model effect is closer to the actual complex environment.

**4.3 Experimental results on the CASIA B dataset**

To evaluate the effectiveness of the SMViT model and the view movement method (SMViT_T) proposed in this paper. We decided to use the first 10,000 training loss changes of the two intermediate views as an assessment of the convergence effect. It can be seen that even in the middle of the view offset, the SMViT model proposed in this paper can still effectively converge and stabilize under the general trend, as shown in the blue line in Figure 4 and Figure 5. The model, after the perspective gradually moves training, not only has the drop in loss, it has been significantly improved, but the unstable jumping phenomenon of losses has also been suppressed to a certain extent. as shown in the other line in Figure 4 and Figure 5.

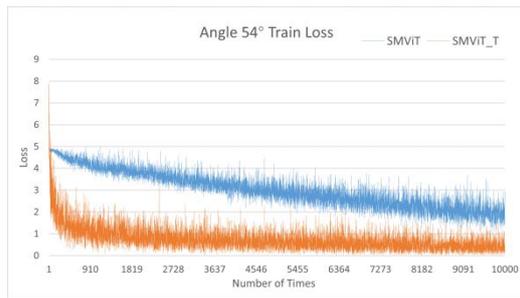
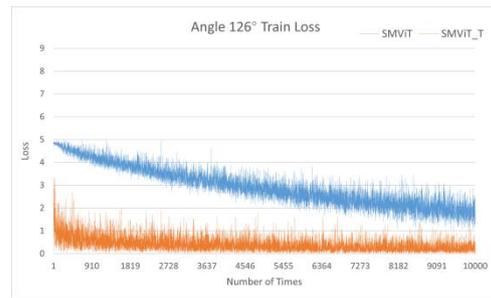

**Figure 4**: Loss change when the view is 54°     **Figure 5**: Loss change when the view is 126°

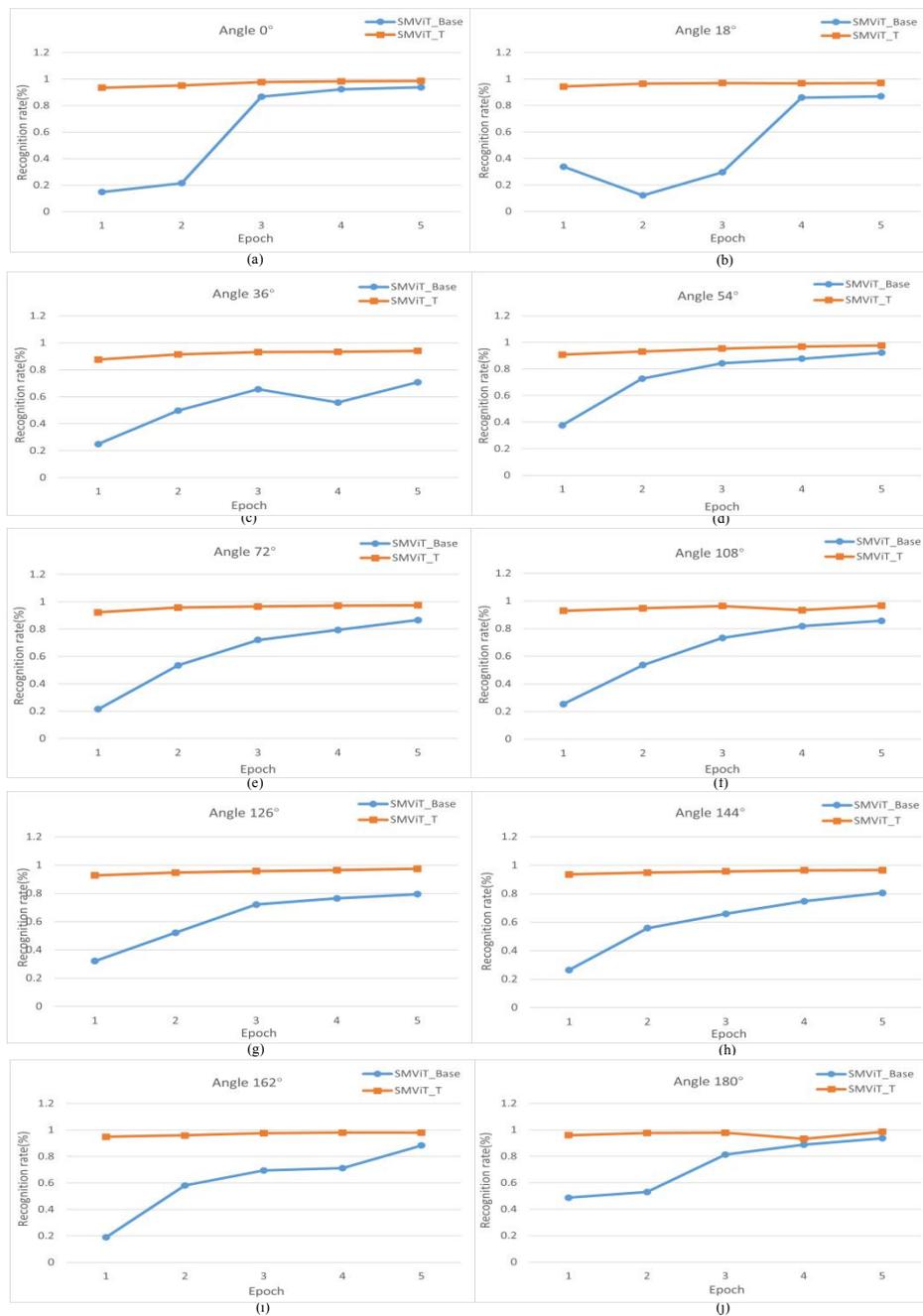

**Figure 6**: Training process diagram of the model in this paper under various views in CASIA B dataset

Figure 6 shows the effect pictures of the model trained with or without a view gradually moving training in 10 views except for 90°. It can be observed that except for the basic model (SMViT_BASE) at 36 ° at the perspective, there is a small accuracy oscillating, and the experimental effects on other views have steadily increased. See the point line in Figure 6. The model in this paper (SMViT_T) with view gradually moving training has high accuracy in all views, and there is no significant fluctuation in the recognition rate during the training process. The recognition rate is always higher than that of the Base model.

**4.4 Ablation Experiment**

**4.4.1 Comparison with the latest technology**

Table 1 Precision comparison of CASIA B with the latest technology in each view

| | Comparison of model accuracy in each view when non-crossing view | | | | | | | | | | |
|---|---|---|---|---|---|---|---|---|---|---|---|
| | 0° | 18° | 36° | 54° | 72° | 90° | 108° | 126° | 144° | 162° | 180° |
| SPAE[23] | 0.7419 | 0.7661 | 0.7150 | 0.6989 | 0.7311 | 0.6801 | 0.6854 | 0.7258 | 0.7016 | 0.6881 | 0.7231 |
| GaitGANv1[11] | 0.6828 | 0.7123 | 0.7285 | 0.7339 | 0.6962 | 0.7043 | 0.7150 | 0.7285 | 0.7204 | 0.7042 | 0.6828 |
| GaitGANv2[12] | 0.7258 | 0.7554 | 0.7150 | 0.7332 | 0.7527 | 0.707 | 0.6962 | 0.7392 | 0.7150 | 0.7311 | 0.6989 |
| Multi_View GAN[24] | 0.7213 | 0.7869 | 0.7814 | 0.7589 | 0.7568 | 0.7131 | 0.7322 | 0.7431 | 0.7431 | 0.7480 | 0.7513 |
| Slack Allocation GAN [25] | 0.7473 | 0.7258 | 0.7258 | 0.7141 | 0.7560 | 0.7336 | 0.6967 | 0.7365 | 0.7277 | 0.7243 | 0.7221 |
| GAN based on U-Net [26] | 0.7365 | 0.7715 | 0.7956 | 0.7957 | 0.8521 | 0.7822 | 0.8172 | 0.7956 | 0.7984 | 0.7419 | 0.7580 |
| PoseGait[27] | 0.7231 | 0.7365 | 0.7688 | 0.7822 | 0.7446 | 0.7473 | 0.7607 | 0.7284 | 0.7553 | 0.7365 | 0.6586 |
| **SMViT_Base** | **0.9802** | **0.9704** | **0.9318** | **0.9805** | **0.9689** | **0.9744** | **0.9668** | **0.9617** | **0.9529** | **0.9451** | **0.9831** |

In the 11 views in CASIA B, the model (SMViT_Base) in this paper is compared with SPAE[31], GaitGAN[19,20], Multi_View GAN[32], Slack Allocation GAN[33], GAN based on U-Net[34], PoseGait[35] in the internal recognition rate of non-cross-view offset view. That is, without considering the cross-verification, verification experiments only in the multi-view. Other model data in Table 1 are taken from the average value of the three-state gait recognition rates from the same view as the same data set. It can be seen that within all views, the model of this paper has greatly improved compared to other generating gait recognition models. And the average upgrade index even exceeds 20 percentage points. The model of this paper is verified that in the multi-view recognition task of non-crossing view, this paper is better than the selected comparison model.

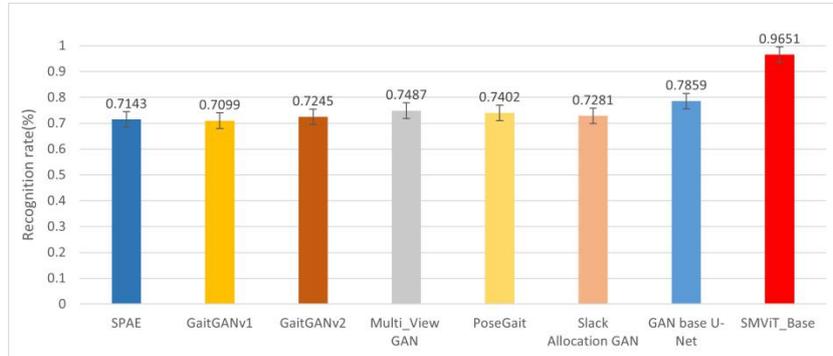

**Figure 7**: Comparison of the average validation rate of the model from multiple views

At the same time, the average value of normal walking (NM), backpack walking (BG), and wearing a jacket (CL) of 11 views is compared. Among them, the training sets and verification sets of each model are taken from the same view. The proportion of internal training sets and verification sets of the view is 7: 3, and the cross-verification of the view is not considered. From Figure 7, it can be seen that the model of red in this paper has also increased significantly in the average value of multi-view mixed recognition rates, which also increased by about 20 percentage points.

**4.4.2 Validation of the gradually moving view training strategy**

Taking the 18 ° gait data as an example to conduct ablation experiments to verify the effectiveness of the view gradually moving training, As can be seen from the square point line in Figure 8, during the first 15 rounds of training, The model SMViT_base in this paper has dropped significantly in the second and 9th rounds, and there is a significant saturation of recognition rates. After experimental analysis, we believe that the first decline was due to losing a lot of gait characteristics outlines，At the same time, this model does not use view mobilization training methods to convert and strengthen the gait characteristics, making the model unable to effectively learn the characteristics, and the accuracy decreases sharply. The second drop is due to the overly small number of features, which leads to the abnormal situation of overfitting gradually in the verification accuracy, which also reflects the improvement of the model robustness by the view transition training method. At around 13 rounds, Base reaches the upper limit of saturation accuracy but is still about one percentage point lower than SMViT_T's recognition accuracy. On the whole, after the gradually moving view training strategy of SMViT, not only is the initial recognition rate of about 70 percentage points higher than the basic model but also maintains a relatively stable recognition accuracy. Although the accuracy saturation trend also appeared soon, the upper limit of saturation value was about one percentage point higher than that of the Base model, and the oscillation amplitude of the validation rate remained below one percentage point.

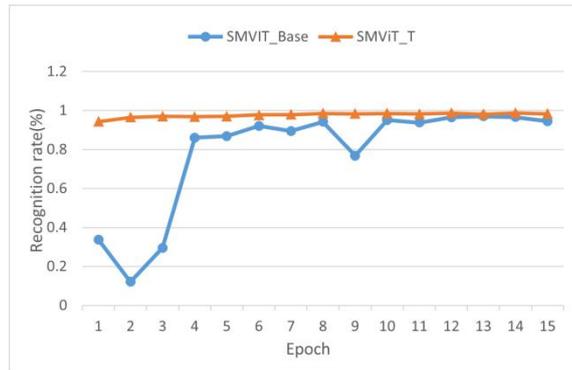

**Figure 8**: Reliability Verification of view gradually moving training at 18 °

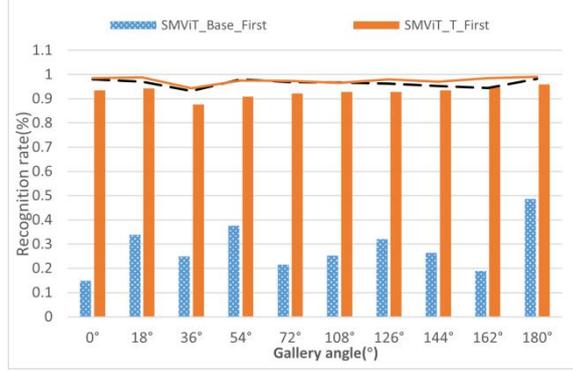

**Figure 9**: Comparison of initial validation rate and maximum validation rate of the proposed model trained with or without view gradually moving training

In order to more intuitively verify the improvement of the validation rate and the stabilization effect of the model by the view of gradually moving training, a set of ablation experiments are designed, as shown in Figure 9. Of the 10 views other than 90° compare the initial validation rate and the maximum validation rate of the model. The model's initial recognition rate in this paper is significantly improved by an average of about 64.4 percentage points using the view gradually moving training, as shown in the bar chart in Figure 9. At the same time, in terms of the maximum recognition rate, the SMViT can be improved to a certain extent and more robust at all views after the view gradually moving training, as shown in the broken line in the figure.

# 5 Conclusion and Prospect

In this paper, we integrate the design thought of Siamese neural networks and a variant Mobile Vision Transformer model and build a multi-view Siamese ViT gait recognition model-SMViT. In the meantime, a view gradually moving training strategy for multi-view gait recognition is designed, called SMViT_Base and SMViT_T, respectively. After conducting a large number of experiments on the CASIA B dataset, it is shown that the method of using the Siamese feature relationship calculation method to obtain the perspective characteristic conversion factor can be used to obtain the relationship between different perspective gait characteristics, which effectively improves the accuracy of multi-perspective step recognition. Experimental results show that the proposed model can significantly improve the recognition rate compared with the existing generative multi-view gait recognition methods without considering cross-view verification. About 20 percentage points increase in the hybrid recognition rate without considering the external dress of pedestrians. Therefore, SMViT expands the gait recognition view while ensuring high accuracy, improving efficient gait recognition in multi-view practical application scenarios.

In the future, a more abundant dataset can be used to verify the recognition effect, and a more sophisticated view feature conversion module can be used to enhance the application scope of SMViT. On the side, when the visible light intensity is insufficient, the infrared thermal imaging target tracking method can also be used to extract gait contour features to deal with more complex and variable natural environments [36] and more hidden gait target tracking tasks [37,38]. We believe that the design of SMViT with multiple covariates will open up new ideas for multi-view gait recognition and The Vision Transformer can also occupy a place in multi-view complex environment gait recognition tasks.